\documentclass[letterpaper, 10 pt, conference]{ieeeconf}
\IEEEoverridecommandlockouts
\usepackage{amsmath,amssymb,amsfonts}
\usepackage{algorithmic}
\usepackage{graphicx}
\usepackage{textcomp}
\usepackage{xcolor}
\usepackage{varwidth}
\usepackage{subfigure}
\usepackage{cite}
\usepackage{comment}
\usepackage{stfloats}

\begin{document}





\title{\LARGE \bf LucidGrasp: Robotic Framework for Autonomous Manipulation of Laboratory Equipment with Different Degrees of Transparency 

via 6D Pose Estimation}

\author{Maria Makarova$^{1}$, Daria Trinitatova$^{1}$, Qian Liu$^{2}$ and Dzmitry Tsetserukou$^{1}$
\thanks{$^{1}$The authors are with the Intelligent Space Robotics Laboratory, Center for Digital Engineering, Skolkovo Institute of Science and Technology (Skoltech), 121205 Moscow, Russia. 
{\tt \small $\{$maria.makarova2, daria.trinitatova, d.tsetserukou$\}$@skoltech.ru}}
\thanks{$^{2}$ Qian Liu is with the Department of Computer
Science and Technology, Dalian University of Technology, China. {\tt\small qianliu@dlut.edu.cn}}
}

\maketitle

\begin{abstract}
Many modern robotic systems operate autonomously, however they often lack the ability to accurately analyze the environment and adapt to changing external conditions, while teleoperation systems often require special operator skills. In the field of laboratory automation, the number of automated processes is growing, however such systems are usually developed to perform specific tasks. In addition, many of the objects used in this field are transparent, making it difficult to analyze them using visual channels.
The contributions of this work include the development of a robotic framework with autonomous mode for manipulating liquid-filled objects with different degrees of transparency in complex pose combinations.  The conducted experiments demonstrated the robustness of the designed visual perception system to accurately estimate object poses for autonomous manipulation, and confirmed the performance of the algorithms in dexterous operations such as liquid dispensing. The proposed robotic framework can be applied for laboratory automation, since it allows solving the problem of performing non-trivial manipulation tasks with the analysis of object poses of varying degrees of transparency and liquid levels, requiring high accuracy and repeatability.
\end{abstract}

 
\section{INTRODUCTION}

Nowadays, the tendency to introduce modern robotic systems into various areas of human activity is undeniable. Apart from fields such as industry, medicine and logistics, robots are increasingly being used in the field of scientific research, especially in various laboratories such as chemical, medical, biological, etc.

Robotic systems can be classified into three main categories based on the type of control, namely teleoperated, shared-control and automated. 
An effective teleoperation system requires an intuitive and robust control interface, as well as a stable and convenient visual feedback channel that allows the operator to quickly respond and adapt to changing conditions of the dynamic environment. Intuitive control methods can be implemented by augmenting the operator with haptic feedback \cite{case2022exploring} and using VR-based interfaces with the robot's digital twin, augmented either by video streaming with object point clouds or visualization of high-fidelity models in recognized poses \cite{naceri2021vicarios, ponomareva2021grasplook}.

Recently, there is a tendency to apply teleoperation interfaces as an effective tool for collecting data on environmental states and operator control signals for subsequent training of a robotic system for autonomous actions using Imitation Learning. Thus, Gello \cite{wu2023gello} and ALOHA \cite{zhao2023learning} systems serve as examples of low-cost and intuitive solutions that exploit kinematic similarities between target robotic manipulators and control interfaces. Despite efficient application of teleoperated robotic systems in different domains, it is difficult to provide precise execution of some labor tasks that require specific operator skills to collect high quality expert data. In some cases, it is more appropriate to build a system with closed-loop robot control algorithms. However, it is required a high-precision perception of the state of the external environment and preliminary verification of actions using a digital twin in a simulated environment, as it was proposed in the current work.

During teleoperation, an operator experiences high workload when solving complex tasks in dynamic environments. To reduce the workload and improve efficiency, various shared-control architectures have been proposed to assist the operator during teleoperation tasks \cite{dop11,dop14}. The results of these studies showed that as the complexity of the task increased, operators preferred not to interfere with the autonomous control of the robot. Considering the manipulation of such complex and often fragile objects, as in the medical or chemical industries, these studies symbolize the need to implement the autonomous system to improve the quality and speed of operations.

Autonomous robotic systems require a detailed closed-loop sensorimotor control system, often based on visual-tactile perception. 
The implementation of automation technologies in the laboratory applications is extremely useful in achieving reproducibility in scientific research and reducing the risks \cite{kitney2019enabling}. In the field of laboratory automation, several systems have been developed. For example, automated processes for solubility determination and crystallization have been proposed by Fakhruldeen et al.\cite{ARChemist}, however the platform architecture requires clearly defined instructions from the operator. The system presented by Lunt et al.\cite{Lunt23} focuses on automating the complex process of powder X-ray diffraction, which is a key technique in materials science and chemistry. As in the aforementioned work, the system lacks a mechanism for analyzing the environment and, as a result, lacks variability of action in dynamically changing external conditions. Nevertheless, a number of automated systems have been proposed that are able to analyze the environment using computer vision. However, these systems are mostly focused on a specific task, such as picking and placing test tubes\cite{Wan21} or solubility screening \cite{Shiri21}.

It should also be noted that transparent and translucent objects are often used in the field of laboratory applications, which complicates their processing using conventional computer vision algorithms. Therefore, we have developed a system for robotic manipulations capable not only of estimating the state of the environment with high accuracy, but also of planning an efficient trajectory for different types of tasks and validating it in a simulated environment with a digital twin. The main contributions of the presented work can be summarized as follows:
\begin{itemize}
    \item Development of an autonomous robotic system for real-time dexterous manipulation of laboratory equipment. The remote environment is analyzed by predicting 6D poses of objects with different degrees of transparency, levels of internal liquids and geometric location of the upper neck of vessels.
    \item The system allows performing a wide range of manipulation tasks based on the algorithmic assignment of only a few key points of the trajectory. In addition, a simulation environment with a digital twin of the robot is used to validate the calculated actions and render the recognized objects from the real environment. 
    \item Experimental verification of the accuracy of the developed visual perception system in a series of experiments and determination of the working area.
\end{itemize}

\section{FRAMEWORK OVERVIEW}

\begin{figure}[!t]
  \centering
\includegraphics[width=0.98\linewidth]{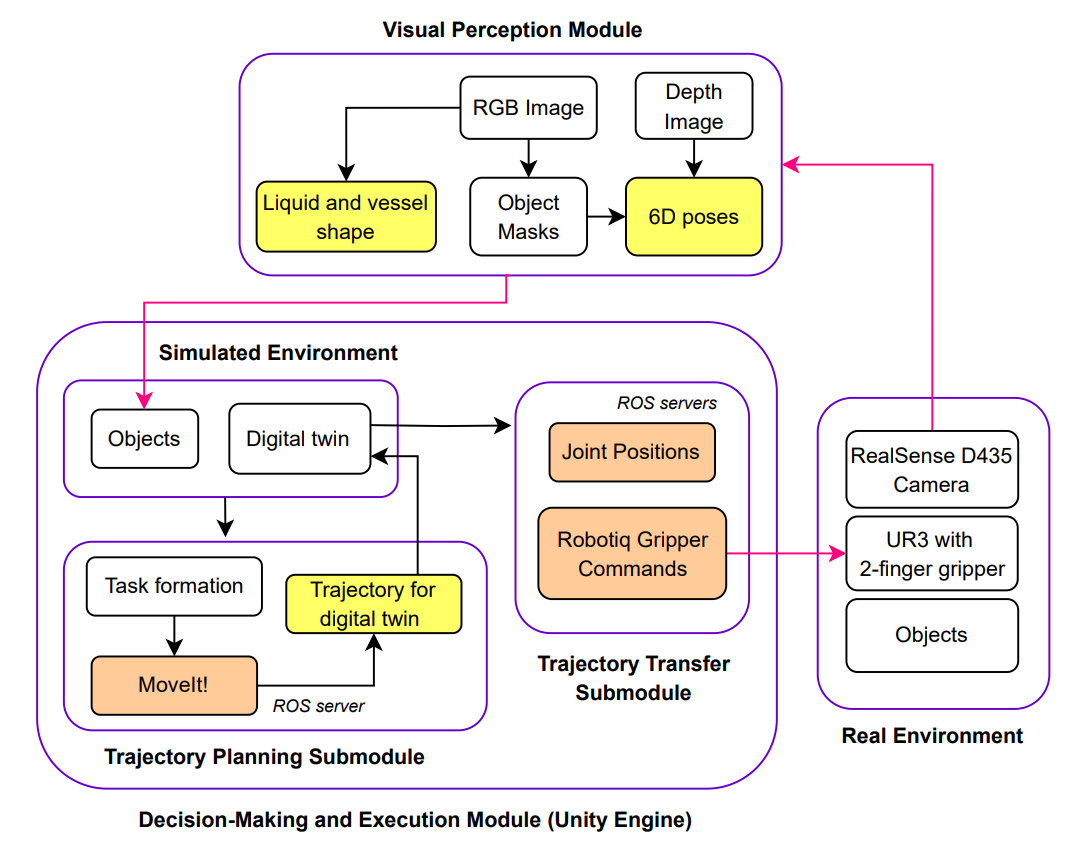}
  \caption{Overview of the proposed robotic framework. The outputs of the main modules are highlighted in yellow, ROS servers in orange.}\label{fig:architecture}
\end{figure}

The scheme of the developed system is shown in Fig. \ref{fig:architecture}. The proposed system is able to detect 6D poses of objects, including translucent and transparent ones. In addition, it can analyze the liquid level in transparent vessels and the geometric location of the neck of the vessels, as well as perform various actions with multiple objects autonomously. It should be noted that the framework allows autonomous execution of a wide variety of manipulation tasks with different configurations of object locations, from object grasping to dispensing operations. The speed and gripping force of the dispenser can also be varied. 

\subsection{System Architecture}

The system includes the \textit{Visual Perception Module}, the \textit{Decision-Making and Execution Module}. The latter consists of the \textit{Simulated Environment}, \textit{Trajectory Generation} and \textit{Trajectory Transfer Submodules}. The framework interacts with the \textit{Real Environment} in which the robotic manipulator (Universal Robots UR3) is located and starts with the operation of the \textit{Visual Perception Module}. It estimates the 6D poses of the objects, as well as the liquid level and geometric location of the vessel neck, using RGB and depth images of objects from the robot's environment (\textit{Real Environment}) captured by the RealSense D435 camera. This information is then used to generate an object manipulation task, after which the trajectory of the robot's digital twin is calculated using the MoveIt!\footnote{https://moveit.ai/} server (\textit{Trajectory Planning Submodule}). This trajectory is transferred to the \textit{Simulated Environment}, where the digital twin manipulates rendered objects whose positions and orientations have been obtained from the \textit{Visual Perception Module}. The values of the robot's joint positions, as well as commands for the two-finger gripper (Robotiq 2F-85) are translated via ROS to a real robotic manipulator (UR3) (\textit{Trajectory Transfer Submodule}). The\textit{ Trajectory Planning Submodule}, the \textit{Trajectory Transfer Submodule}, and the \textit{Simulated Environment} are implemented using the Unity engine and integrated into a large module called the \textit{Decision-Making and Execution Module}.
 
\subsection{Visual Perception Module}
The \textit{Visual Perception Module} receives RGB and depth images of the \textit{Real Environment} using the RealSense D435 camera. The RGB image is used to predict object segmentation masks as well as liquid and vessel shapes. The depth image and segmentation masks are used to obtain the 6D poses of objects. 

\subsubsection{6D Pose Estimation}
Currently, a variety of different model architectures have been proposed to predict the 6D poses of objects. For example, it is possible to match 3D models to observed objects using direct regression, but this approach becomes resource-consuming as the number of instances increases \cite{he2021ffb6d}. Wang et al. \cite{wang2019densefusion} presented the DenseFusion architecture, which allows building a single model for multiple objects. However, this requires expensive re-training every time a new object instance is added to the database. Park et al.\cite{park2020latentfusion} proposed the LatentFusion framework, which reconstructs a latent 3D object model from a small set of reference views, and later infers the 6D pose from the input image. The proposed approach is computationally expensive since it is based on iterative optimization at inference time. 

In the current work, 6D object pose estimation is performed from a single depth image and the object segmentation mask using the OVE6D architecture \cite{cai2022ove6d}, which generalizes to new objects without any re-training of model parameters. In addition, the applied model is computationally efficient and robust to occlusions in the input data. The OVE6D architecture utilizes the viewpoints of the object obtained by its renderer. The viewpoints are computed using 3D mesh models of objects that are preloaded into the neural network. The model is able to sequentially predict the position of the object from the given viewpoint, and then add in-plane orientation regression to the desired angle. This allows the architecture to be robust and produce sufficient pose estimation results even for transparent objects. At this stage of development, no additional training of the OVE6D architecture on the custom dataset was required. 
 
\begin{figure}[!t]
  \centering
 \subfigure{\includegraphics[width=0.47\linewidth]{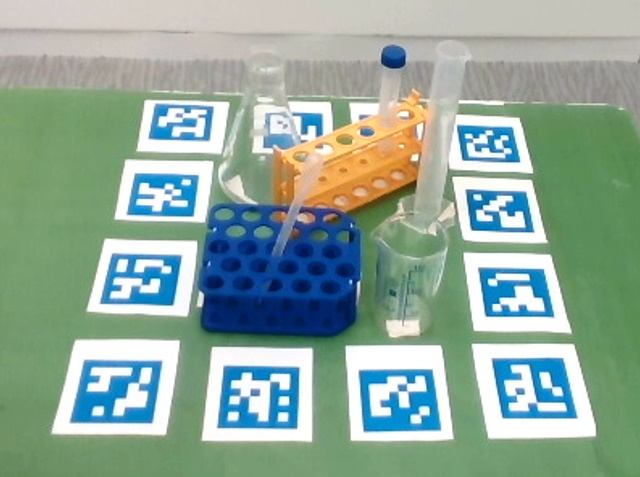}}
  \subfigure{\includegraphics[width=0.435\linewidth]{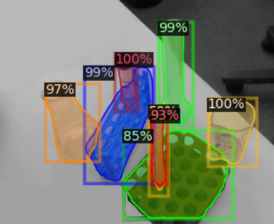}} 
  \caption{Dataset objects (left) and object segmentation masks (right).}\label{fig:target_obj}
\end{figure}

The target objects for manipulation are 7 different objects of laboratory equipment (test tube, pipette, glass beaker, volumetric flask, graduated cylinder, and two tube racks), many of which are transparent or translucent. 3D mesh models of each of these are also available for the model. Due to the transparency of the objects, 6D pose recognition from RGB and depth images becomes a challenging task compared to the recognition of opaque objects. The segmentation masks fed to OVE6D were obtained using Mask R-CNN \cite{he2018mask}, which was trained on a collected video dataset consisting of 2500 images (Fig. \ref{fig:target_obj}). The dataset was collected using Intel RealSense D435 camera in the format of LINEMOD benchmark\footnote{https://bop.felk.cvut.cz/datasets/} that contains RGB and depth images with segmentation masks and 3D object mesh models. It should be mentioned that ARUCO markers\cite{aruco} are not used in the framework algorithms and were only needed during the data collection phase. An example of the output segmentation masks is shown in Fig. \ref{fig:target_obj}. It is planned to eliminate the need to train segmentation models in future work, for example by using Vision Language Models (VLMs). 

To transfer the estimated poses to the \textit{Simulated Environment} (Unity Engine), the obtained coordinates and angles are recalculated in the camera coordinate system. The new coordinate reference point is located at the base of the robot. This data is then transferred to the \textit{Trajectory Planning Submodule} and the \textit{Simulated Environment} via TCP/IP. 

\subsubsection{Predicting the Liquid Shape and Vessel Neck}

\begin{figure}[!ht]
  \centering
  {\includegraphics[width=0.98\linewidth]{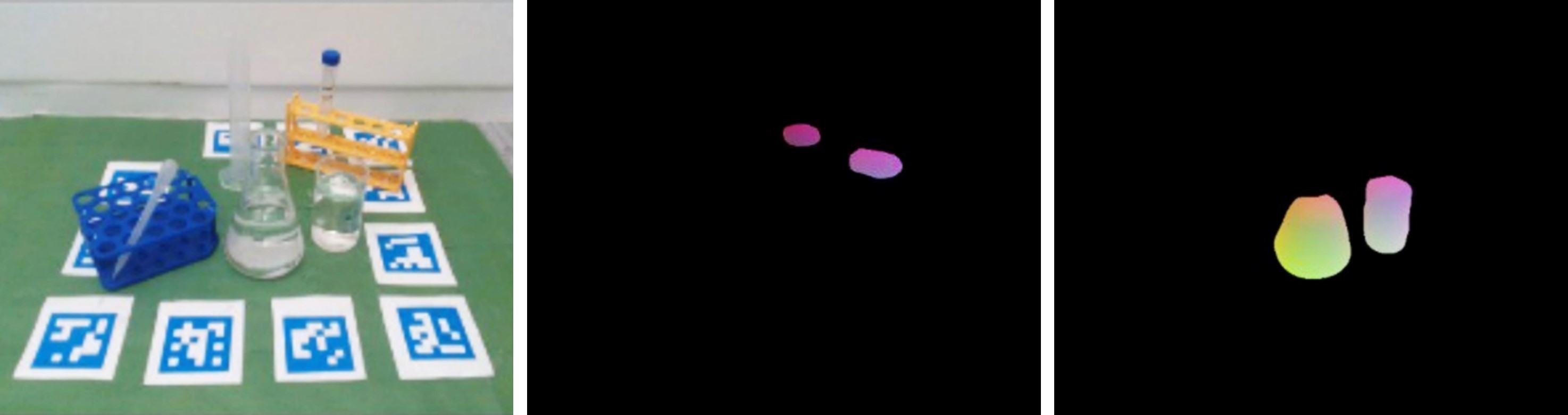}}   
  \caption{Example of predicting the neck of a vessel and the shape of the liquid inside the vessel in an occluded environment.}\label{fig:liquid_pred}
\end{figure}

To operate autonomously, the system requires the ability to independently create the task conditions for manipulation. The proposed system is able to recognize the shape of the liquid in transparent objects and the geometric location of the vessel neck points, which helps to calculate the robot's trajectory for proper operation.  Using the method proposed by Eppel et al. \cite{eppel2022predicting},  we applied a model consisting of a fully convolutional network \cite{long2015fully} with an atrous spatial pyramid pooling (ASPP) dilated convolutional decoder \cite{chen2017rethinking}, a Resnet101 encoder \cite{he2015deep}, and three layers of skip connection and upsampling \cite{ronneberger2015unet}. Prediction of the vessel, vessel content and vessel neck maps formed the final layer of the applied network (Fig. \ref{fig:liquid_pred}). Similar to the OVE6D model, no additional training on the custom dataset was required. 

\subsection{Decision-Making and Execution Module}\label{digital twin}
As shown in Fig. \ref{fig:architecture}, the \textit{Decision-Making and Execution Module}, which contains all the operational logic, consists of the \textit{Trajectory Planning Submodule}, the \textit{Trajectory Transfer Submodule} and the \textit{Simulated Environment}.

\subsubsection{Simulated Environment} \label{sim_env}
The \textit{Simulated Environment} is based on the Unity engine and consists of a digital twin of the robotic manipulator (UR3) and objects whose poses are updated according to information from the \textit{Visual Perception Module}. Once the trajectory has been calculated using the \textit{Trajectory Planning Submodule}, the digital twin of the robot executes the necessary commands in a \textit{Simulated Environment} with rendered objects. This allows the algorithms to be validated before they are executed by the real robot. 

\subsubsection{Trajectory Planning Submodule}
Firstly, the trajectory of the robot's digital twin is planned based on the location of the objects and the type of task. This is implemented in the \textit{Trajectory Planning Submodule}. After analyzing the position of the objects, only a few key points of the trajectory are required to be algorithmically calculated to plan the robot motion (\textit{Task Formation}). For the object picking task, there are Pre-Grab, Grab, Pick, Place and PostPlace robot poses. For the more complex tasks discussed in section \ref{autonomus_dem}, robot poses are computed iteratively for each object simultaneously with the gripper commands. All these poses are translated to the MoveIt! server to plan and actuate the robot actions for the digital twin in the \textit{Simulated Environment} (Fig. \ref{fig:architecture}).

\subsubsection{Trajectory Transfer Submodule}
During the task execution by the digital twin, the robot's joint positions as well as commands for the Robotiq gripper are translated to the real robot with the help of several ROS servers\footnote{https://www.ros.org/}. This is implemented in the \textit{Trajectory Transfer Submodule}. It operates in real-time mode, which is ensured by simultaneously running ROS servers and communicating with them through separate types of ROS messages generated for each task.

\section{EXPERIMENTS}\label{expirements}

\subsection{Defining Work Area for 6D-pose Estimation}
\subsubsection{Changing camera height at a fixed distance}
Since the pose detection errors increase with decreasing distance between the camera and the objects, the performance of the algorithm was analyzed at a close distance of 9.5 $cm$ horizontally from the nearest edge of the board with objects. Three different values of camera height above the board were considered, namely 50, 45 and 40 $cm$. At each height, the value of the camera angle of view along the pitch axis was varied three times between 40$^\circ$ and 65$^\circ$ in 5$^\circ$ increments. For each height, the angles were chosen so that all objects were clearly visible over the whole area.

\begin{figure}[!h]
  \centering
  {\includegraphics[width=0.88\linewidth]{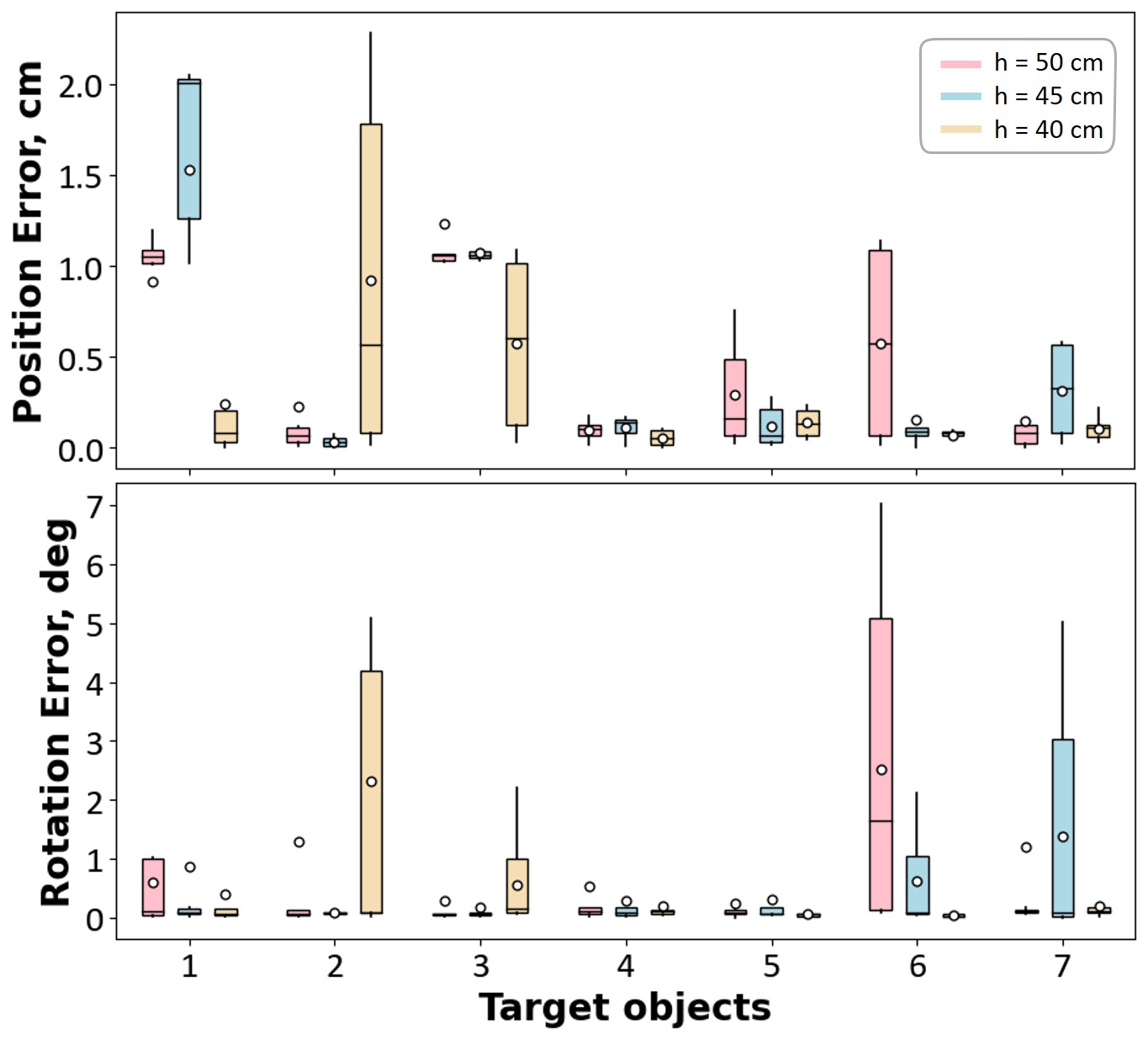}}
  \caption{Position and rotation errors of target objects during pose estimation: 1 -- flask, 2 -- glass beaker, 3 -- graduated cylinder, 4 -- pipette, 5 -- test tube, 6 -- 6-hole tube rack, 7 -- 25-hole tube rack.}\label{fig:errors_height_changing}
  \vspace{-0.5em}
\end{figure}

\textit{Experimental results}:
For each height value, the average errors in position ($x$ and $y$ coordinates) and rotation of each object were estimated. The results obtained in terms of recognition accuracy are presented in Fig. \ref{fig:errors_height_changing}. The horizontal axis shows the indices of each of the seven target objects.

We analyzed the recognition accuracy of target objects, averaged over the camera heights, using Kruskal-Wallis non-parametric test, with a chosen significance level of $\alpha<.05$. According to the test results, there is a statistically significant difference in position recognition accuracy between the target objects $(H = 32.1,p<.001)$ . The position recognition was most unstable for large transparent and translucent objects such as flask, glass beaker and graduated cylinder (objects 1,2 and 3 respectively). Comparing two groups of transparent and translucent objects, we found that the position recognition of smaller translucent objects such as pipette and test tube were statistically significantly better than larger ones $(p=.001)$ according to Mann-Whitney U test. This suggests that the distance to the objects should be increased for stability.
Analyzing the rotation recognition accuracy, it should be noted that the largest error variation was observed for asymmetric objects such as tube racks (objects 6 and 7).
Overall, the minimum errors for pose estimation were obtained at the height of 40 $cm$. Thus, the mean position error averaged over all objects comprised 0.3 $cm$ (SD=0.52 $cm$), while the mean rotation error was 0.54$^\circ$ (SD=1.6$^\circ$).

\subsubsection{Changing Camera Distance with a Fixed Height}

Having analyzed the dependence of pose estimation accuracy on the camera height and viewing angle, we additionally analyzed the algorithm performance at various camera distances from the board with objects. For this experiment, the camera mounting height was fixed. Object poses were analyzed for six distance markers, namely 9.5, 13, 24, 33, 57, 65 and 74 $cm$. Camera rotation angles were chosen to provide the same angle of view in each case.

\begin{figure}[!ht]
  \centering
\includegraphics[width=0.92\linewidth]{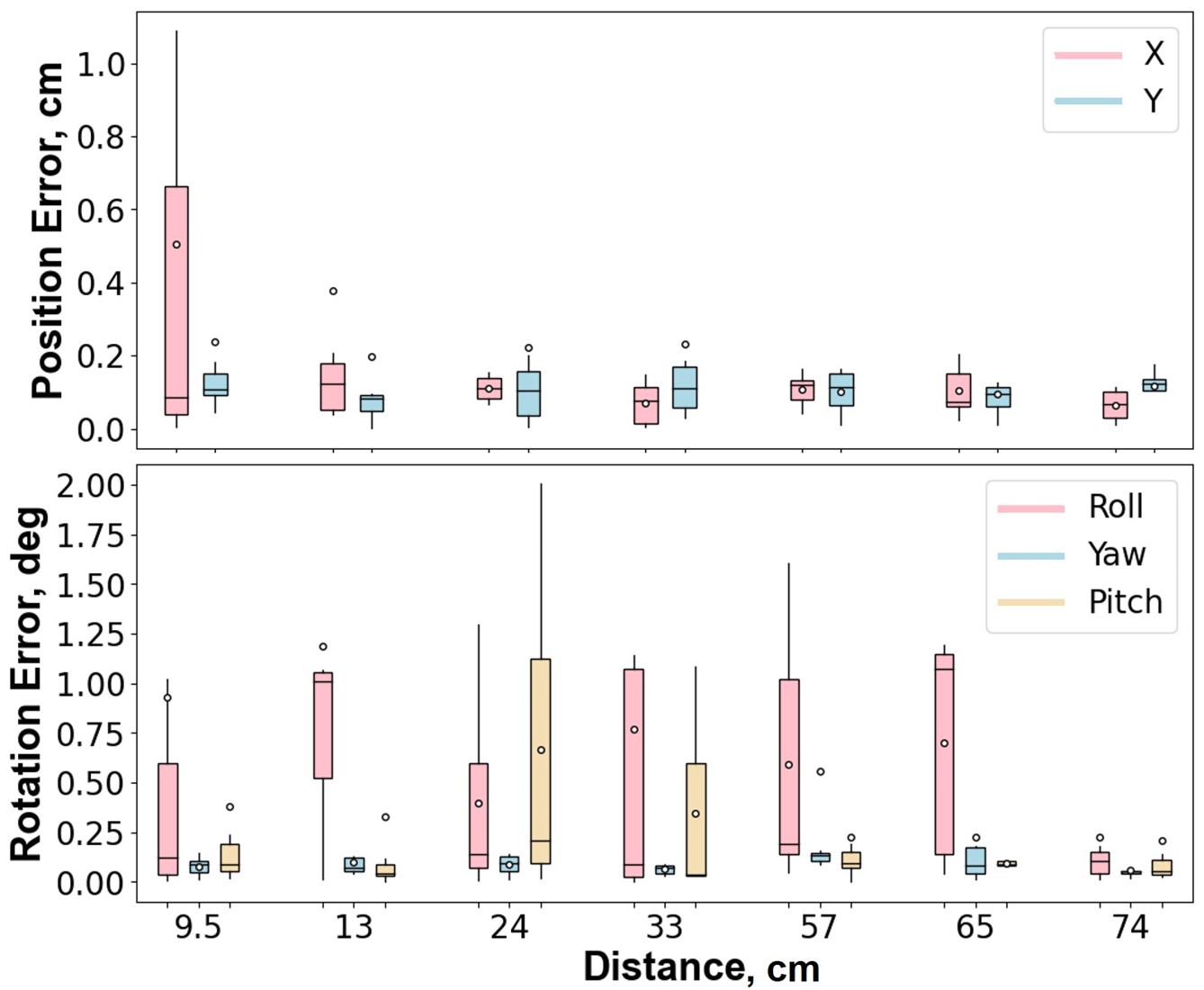}
  \caption{Dependence of position  and rotation errors, averaged over all objects, on the distance of the camera to the objects.}\label{fig:exp_distances}
\end{figure}

\textit{Experimental results}:
The resulting distribution of averaged position and rotation errors is shown in Fig. \ref{fig:exp_distances}.
The experimental results were analyzed using Kruskal-Wallis non-parametric test, with a chosen significance level of $\alpha<.05$, since the obtained data deviated from normal distribution. According to the test findings, there is no statistically significant difference in the pose estimation errors averaged over all objects for different camera distances. The mean absolute position and rotation errors averaged over all objects comprised 0.18 $cm$ and 0.39$^\circ$ respectively. It should be noted that at close distances, the main contribution to the error in recognition along the $X$-axis and recognition of the roll angle was made by a glass beaker, which is a simple cylindrical shape.

In addition, we analyzed the average accuracy of recognition of translucent (objects 1–5) and opaque objects (objects 6–7) using the Mann-Whitney U test for pairwise comparison. According to the obtained results, there is no statistically significant difference in accuracy of recognition for position  $(p=.52)$ and rotation $(p=.56)$ between these two groups of objects. Thus, we can conclude that the recognition of translucent objects was as reliable as the opaque ones.

As a result of the experiments, the boundaries of the working area of the \textit{Visual Perception Module} have been clarified, and the algorithm shows the best results when the camera is located at a medium distance from the objects. When adapting the pose estimation algorithm for another system, the working area should also be determined experimentally.

\subsection{6D-pose Estimation for Complex Object Combinations}
\begin{figure}[!ht]
  \centering
  \subfigure[Examples of complex object poses with occlusions.]{ \includegraphics[width=0.91\linewidth]{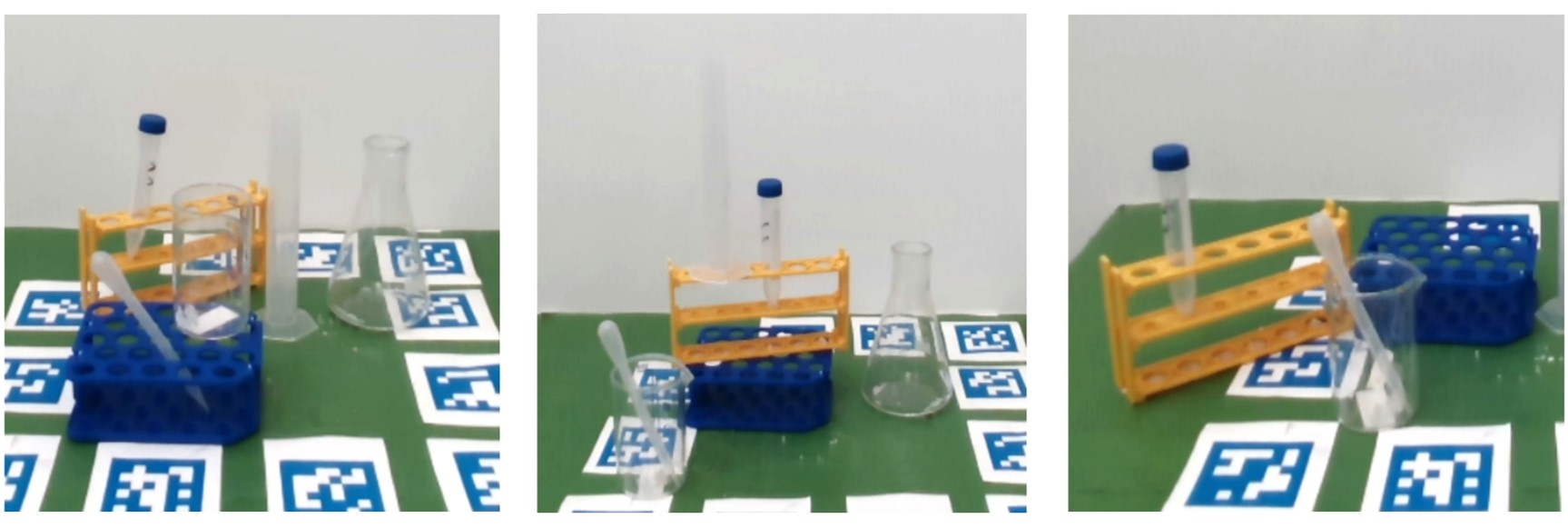}}
   \subfigure[Example of complex object pose estimation.]{\includegraphics[width=0.91\linewidth]{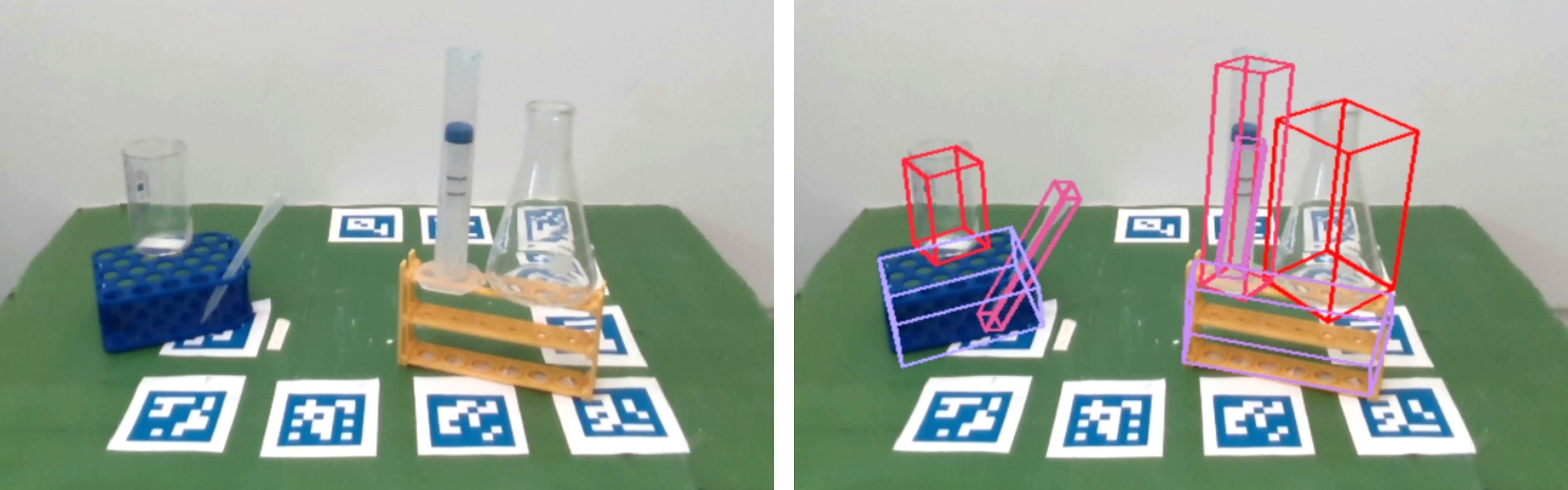}}
  \caption{Complex combinations of objects used in the experiment.}\label{fig:complex_pos}
   \vspace{-0.5em}
\end{figure}

After defining the working range of the pose estimation algorithm, we conducted the experiment to detect poses in random complex combinations of objects.
We estimated cases such as arrangement of objects on top of each other up to four levels in height, the placement of transparent objects on a white background or vice versa on a complex background of opaque objects, as well as the placement of one transparent object inside another (Fig. \ref{fig:complex_pos}).

\begin{figure}[!ht]
  \centering
 \includegraphics[width=0.9\linewidth]{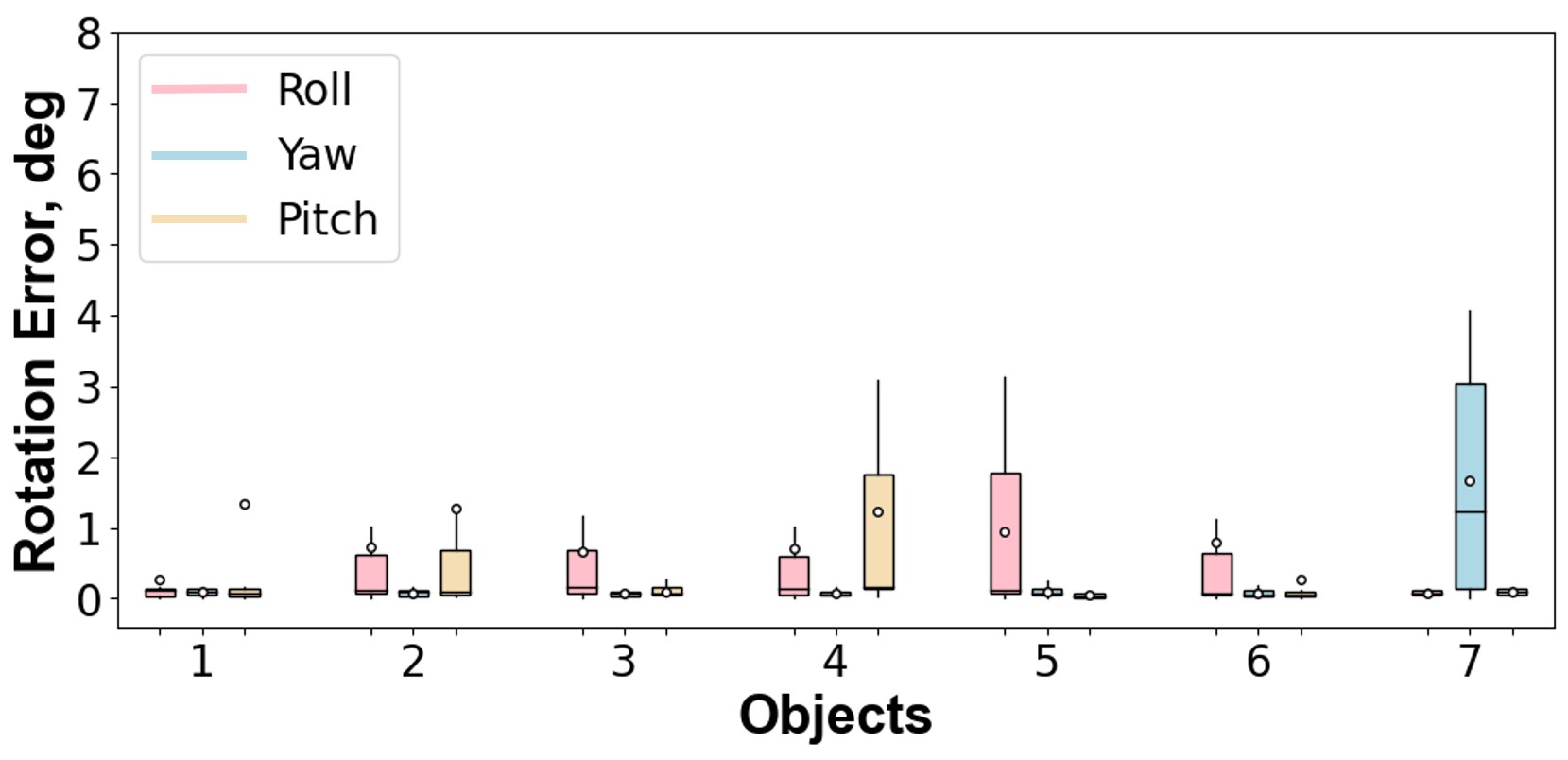}
  \caption{Dependence of rotation errors estimated for complex object combinations. 1 -- flask, 2 -- glass beaker, 3 -- graduated cylinder, 4 -- pipette, 5 -- test tube, 6 -- 6-hole tube rack, 7 -- 25-hole tube rack.}\label{fig:complex_pos_res}
 \vspace{-0.5em}
\end{figure}

The algorithm successfully coped with pose detection cases when both transparent and opaque objects are partially occluded. The average accuracy results obtained are summarized in Fig. \ref{fig:complex_pos_res}. 
We only analyzed rotation errors, since position errors were insignificant. The main detection problems occurred when tube racks were occluded by more than 60\% of the observed surface and when the bottom of the glass beaker was occluded, which, unlike a flask, has parallel walls and therefore a smaller reflective surface. The mean errors for estimation of roll, pitch, and yaw angles averaged over all objects comprised $0.6^\circ (SD=1.1^\circ$),  $1.1^\circ (SD=3.6^\circ$) and  $0.5^\circ (SD=1.5^\circ$), respectively. These results confirm the stable operation of the algorithm for determining 6D-positions even in a complex joint configuration of objects. It is worth mentioning that the accuracy can be improved through fine-tuning the OVE6D model on its own dataset.

\subsection{Demonstration of Autonomous Manipulation in an Occluded Environment}\label{autonomus_dem}

In this experiment, the autonomous robotic manipulation was tested to perform a liquid dispensing operation. The task was formulated as follows: grasp a tilted pipette, draw up liquid with a pipette from the glass beaker, and pour it into a flask.
After exploring the working area of the algorithm in the previous experiments, it was defined the camera location for stable and reliable object recognition. 

\begin{figure}[!h]
  \centering
  \includegraphics[width=0.91\linewidth]{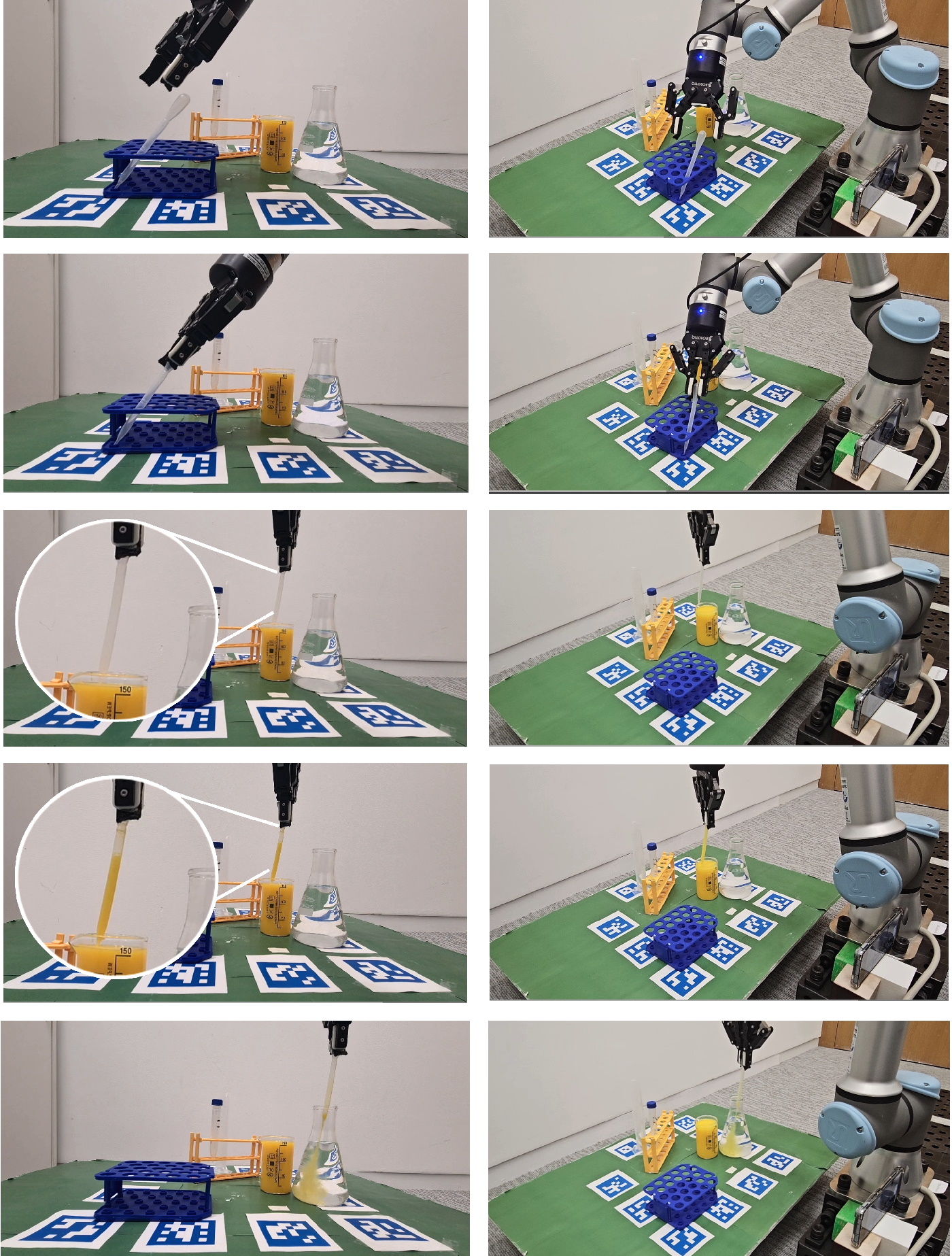}
  \caption{Illustration of performing autonomous liquid dispensing operation. 1 -- Preparing to grip the pipette, 2 -- Grasping the pipette, 3 -- Lowering the pipette into the vessel containing liquid (glass beaker), 4 -- The process of drawing liquid into the pipette, 5 -- Pouring liquid from the pipette into another vessel (flask).}\label{fig: autonomus_system}
\end{figure}
After recognition, the liquid in the first vessel (glass beaker) was painted for better visualization. As described in the section \ref{sim_env}, the robot trajectory calculation is first performed for the digital twin to check all operations in Unity before connecting the real robot to the system. 

The stages of completing the task are shown in Fig. \ref{fig: autonomus_system}. Based on the recognized object poses, six key points of the trajectory were algorithmically calculated.  
The \textit{first} point is the point where the gripper is rotated parallel to the pipette to grasp it, the \textit{second} is the gripping point, the \textit{third} is the point where the gripper with the pipette is rotated perpendicularly over the vessel containing the yellow liquid (glass beaker), the \textit{fourth} is the point where the pipette is lowered into the vessel to draw the liquid, the \textit{fifth} is the point above the second vessel (flask), and the \textit{sixth} is the point where the pipette is lowered into it to pour out the liquid. 
The optimal trajectory between these points was calculated using MoveIt!. All operations were performed accurately and without collisions with other objects. This effect was achieved by algorithmically determining safe positions for lifting the gripper over each object before moving on to the next, and by defining dead zones around each object that are not currently being manipulated. These parameters were calculated from 6D object pose data.

\section{CONCLUSIONS AND FUTURE WORK}

In this work, we have presented a framework for autonomous robotic manipulation of objects with different degrees of transparency.
The proposed system is capable of estimating 6D poses of objects arranged in a variety of location configurations, the level of internal liquid, the geometric location of the upper neck of vessels, and autonomously manipulate objects in various tasks.
In the experimental evaluation, the framework has demonstrated an average accuracy of 0.18 $cm$ for position estimation and about 0.7$^\circ$ for rotation estimation for complex combinations of objects in the algorithm working area. This demonstrates the robustness of the framework's autonomous algorithms.

As a future work, it is planned to use information from tactile sensors on the robotic gripper to control the gripping force more precisely when handling fragile objects. The framework can also be extended with functions that generate key trajectory points for more diverse tasks, as well as by adding VLMs to the system architecture.

The proposed framework can be potentially implemented for the automation of non-trivial tasks of manipulating objects with different degrees of transparency with additional analysis of the liquid level inside, requiring high accuracy and repeatability.  We believe that the capabilities of the developed system may be essential in the field of automated chemical experiments and in medical analysis.

\section{ACKNOWLEDGMENT} 
Research reported in this publication was financially supported by the RSF grant No. 24-41-02039.

\bibliographystyle{IEEEtran}
\bibliography{bib}

\end{document}